\documentclass[11pt]{article}

\usepackage{microtype} 
\usepackage{booktabs}  
\usepackage{graphicx}
\usepackage{amsmath}
\usepackage{amsthm}
\usepackage{makecell}
\usepackage{url}
\usepackage{multirow}

\usepackage[hidesupplement,final]{automl}

\usepackage{natbib}
\title{ALBench: A Framework for Evaluating Active Learning in Object Detection}

\author[1]{\nameemail{Zhanpeng Feng}{zpfeng.cs@gmail.com}}
\author[2]{\nameemail{Shiliang Zhang}{slzhang.jdl@pku.edu.cn}}
\author[2]{\nameemail{Rinyoichi Takezoe}{takezoe@pku.edu.cn}}
\author[1]{\nameemail{Wenze Hu}{windsor.hwu@gmail.com}}
\author[3]{\nameemail{Manmohan Chandraker}{manu.chandraker@gmail.com}}
\author[4]{\nameemail{Li-Jia Li}{lijiali@cs.stanford.edu}}
\author[5]{\nameemail{Vijay K. Narayanan}{vijay_k_narayanan@hotmail.com}}
\author[1]{\nameemail{Xiaoyu Wang}{fanghuaxue@gmail.com}}

\affil[1]{Lighthouse}
\affil[2]{School of Computer Science, Peking University}
\affil[3]{University of California, San Diego}
\affil[4]{Stanford University}
\affil[5]{ServiceNow}

\hypersetup{%
  pdfauthor={}, 
  pdftitle={},
  pdfsubject={},
  pdfkeywords={}
}

\begin{document}

\maketitle

\begin{abstract}
Active learning is an important technology for automated machine learning systems. In contrast to Neural Architecture Search (NAS) which aims at automating neural network architecture design, active learning aims at automating training data selection. It is especially critical for training a long-tailed task, in which positive samples are sparsely distributed. Active learning alleviates the expensive data annotation issue through incrementally training models powered with efficient data selection. Instead of annotating all unlabeled samples, it iteratively selects and annotates the most valuable samples. Active learning has been popular in image classification, but has not been fully explored in object detection. Most of current approaches on object detection are evaluated with different settings, making it difficult to fairly compare their performance. To facilitate the research in this field, this paper contributes an active learning benchmark framework named as ALBench for evaluating active learning in object detection. Developed on an automatic deep model training system, this ALBench framework is easy-to-use, compatible with different active learning algorithms, and ensures the same training and testing protocols. We hope this automated benchmark system help researchers to easily reproduce literature's performance and have objective comparisons with prior arts. The code will be release through Github\footnote{https://github.com/modelai/ALBench}.
\end{abstract}

\section{Introduction}
An automated machine learning system involves multiple components, including automated model design, training data selection, \emph{etc}. Currently, a large body of works focus on the model design. Related works include neural architecture search, meta-learning, and hyper-parameter optimization, \emph{etc.} The research on the other component, \emph{i.e.}, automated data selection has not been as intensive. It is partially attributed to the complication of performance evaluation. The resulting model output by a model design strategy can be easily evaluated using standard classification/detection/segmentation datasets, following commonly used training/testing paradigms without any ambiguity. Differently, the data automation process could involve sampling strategies with extra setups, such as the number of initial training samples, the number of samples added in each iteration, when to stop, \emph{etc.} Unfortunately, different works propose different practices which makes direct apple-to-apple comparison difficult. It obviously does not benefit the prosperity of the community.

Active learning is one of the most important techniques for data selection automation. It~\citep{kovashka17, Li2007, Grauman2014} has been proposed to address the challenge of expensive data annotation encountered by AI models, especially deep models~\citep{he2016deep} which could consume millions or even billions of samples. Instead of annotating every sample available and training the model all at once, active learning trains models in multiple steps. Each step tends to select and annotate a small portion of samples, which are most valuable for subsequent performance enhancement. Combined with incremental learning~\citep{Kading2017}, active learning has shown great potentials in image classification, \emph{e.g.}, it significantly reduced the annotation cost (60\%) while achieving performance on par with that of using full data~\citep{Caramalau2021}. Given the fact that academic datasets used for evaluating active learning have been dedicatedly prepared with sample selection and noise filtering, the improvement from active learning could be even more substantial for open world scenarios, which commonly contain more noises and redundancies.

\begin{table}[t] \small
\setlength{\tabcolsep}{1.6mm}{
\footnotesize
\begin{center}
\begin{tabular}{l|p{1in}|c|c|c|c}
\hline
Method & Training Set & Initial Size & Added Size & Stop Size & Testset\\
\hline
~\cite{Aghdam2019} & \makecell{Caltech Pedestrian, \\ CityPersons, \\ BDD100K}  & 500  & 500 & 7.5K & Caltech Pedestrian testset\\
\hline
~\cite{Haussmann2020} & self collected & 100K  & 200K  & 700K & self collected  \\
\hline
~\cite{Yoo2019} & VOC07, VOC12  & 1K  & 1K & 10K & VOC07 testset \\
\hline
  & VOC07, VOC12   &  1K & 1K & 10K & VOC07 testset  \\
~\cite{Choi2021} & VOC07   & 2K  & 1K & 4K & VOC07 testset  \\
 &  COCO14 &  5k & 1k & 7k & COCO17 VAL \\
\hline
  &  VOC12 & 500  & 200 & 3.5K & VOC12 testset \\
~\cite{Kao2018} &  VOC07 & 500  & 200 & 3.5K & VOC07 testset \\
 & COCO14  & 5K  & 1K & 9K & COCO14 val \\
\hline
~\cite{Di2019} &  KITTI & 1K  & 200 & 12K & KITTI self-divided \\
\hline
~\cite{Brust2018} & VOC12  &  50 & 50 & 250 & VOC12 self-divided \\
\hline
& VOC07  & 500  & 500 & 3.5K & VOC07 testset  \\ 
~\cite{Zhu2021}&  VOC12 & 500  & 500 & 3.5K & VOC12 testset  \\
 & COCO14  &  5K & 1K & 9K & COCO14 VAL   \\
\hline
~\cite{Roy2018} & VOC07, VOC12  &  1655 & 827 & 7447 & VOC07 testset \\
\hline
~\cite{Yuan2021} & VOC07, VOC12  & 827 & 413 & 3.31K & VOC07 testset \\
 & COCO14  & 2340 & 2340 & 11.7K & COCO14 VAL \\

\hline

\end{tabular}
\end{center}
}
\caption{Training and testing setups utilized in existing active learning works for object detection, where ``Added Size'' denotes the number of selected samples at each iteration. The training stops when the size of augmented training set reaches the ``Stop Size''. It shows that current methods employ various benchmark setups which prevents objective comparison among them.}
\label{tab:setup-comparison}
\end{table}

Research on active learning for object detection has not been as popular as that for image classification. This is partially because it involves more training setups than traditional learning strategies, making the algorithm implementation, repeat, and comparison difficult. Besides that, existing active learning algorithms for detection follow different evaluation paradigms, making their performance not directly comparable. We summarize the training and test setups in existing works in Table~\ref{tab:setup-comparison}. The listed studies could use:
\begin{itemize}
    \item \textbf{different datasets.} For instance, the algorithm proposed by~\cite{Yoo2019} is only evaluated on the PASCAL VOC~\citep{Everingham2010} dataset, and is not tested on the more challenging COCO dataset~\citep{COCO2014}. Some approaches are tested on autonomous driving datasets, rather than generic object detection datasets. 
    \vspace{-3mm}
    \item \textbf{different initialization settings.} Both~\citet{Yoo2019} and~\citet{Kao2018} perform experiments on the PASCAL VOC dataset, but use different number of samples in their initial training sets, respectively.
    \vspace{-3mm}
    \item \textbf{different sampling settings for each iteration.} The method~\citep{Haussmann2020} selects 200k images at each training step, substantially more than those determined by other approaches~\citep{Yoo2019}. Introducing more samples at each step helps to boost the model performance, but leads to higher annotation cost. 
    \vspace{-3mm}
    \item \textbf{different stop criteria.} Both~\citep{Yuan2021} and~\citep{Zhu2021} are tested on the MS COCO dataset, but those two methods end up with 11.7K and 9K training samples, respectively. Stopping with more data definitely benefits the final performance.
\end{itemize}


\begin{figure}
\centering
\includegraphics[width=0.7\linewidth]{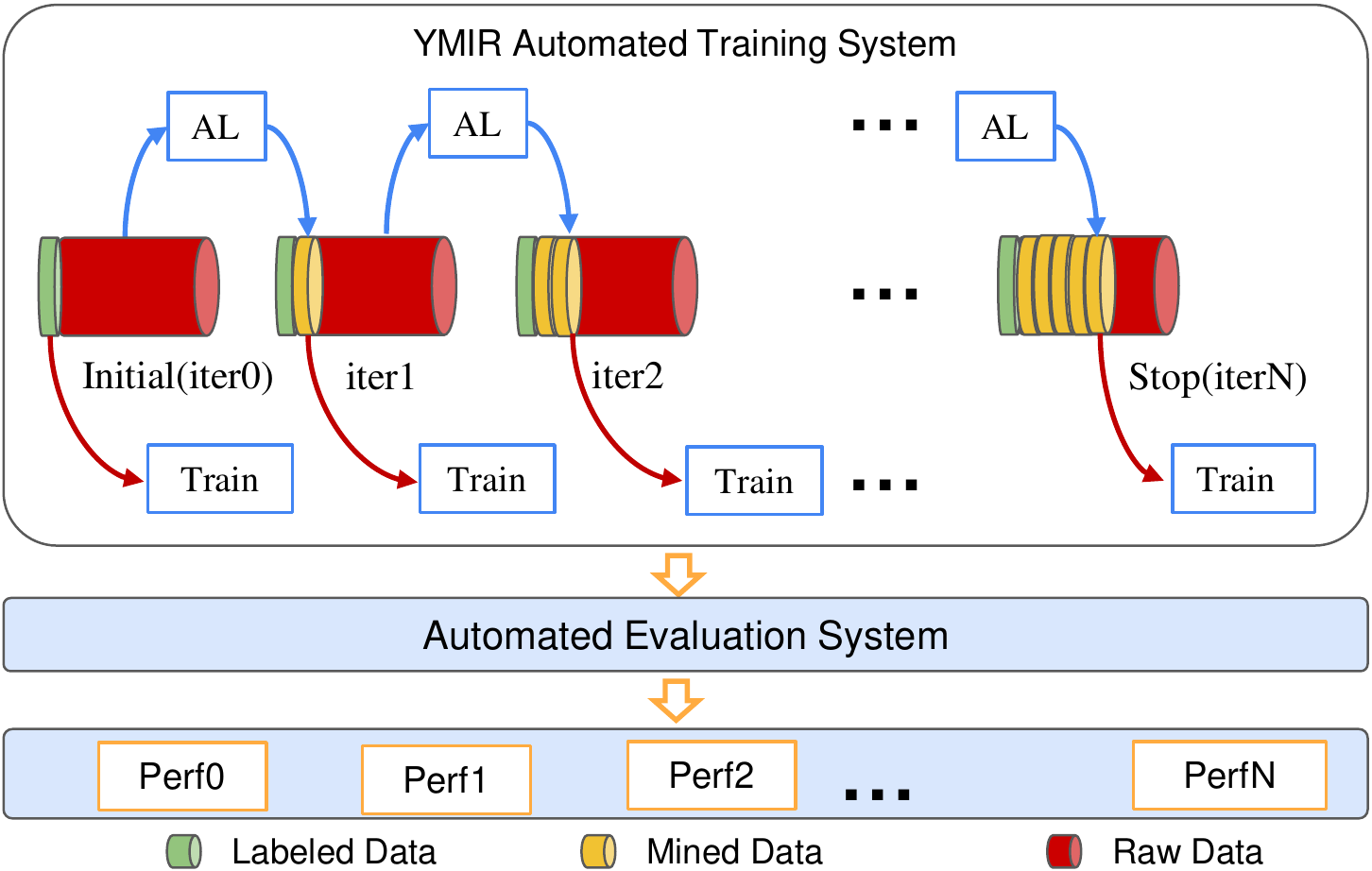}\\
\caption{Illustration of active learning procedure implemented by ALBench based on YMIR~\citep{YMIR2021}, where ``AL'' denotes the active learning model provided by authors. ``Train'' denotes the training action using the augmented training data, each training action typically produces a model. The automated evaluation system collects all the models obtained from training actions in each iteration and produces performance profiling for the corresponding model (Perf0, Perf1, \emph{etc.}). The comprehensive profiling results are used to compare with those of other methods, and generate the leaderboard accordingly. The whole process is automatic once the author provides the training and active learning code for a single iteration.}
\label{fig:alpipeline}
\end{figure}

As the performance of active learning is closely related to both training and test settings, those observations motivated us to construct a benchmark framework for the community that is easy to reproduce and fairly compare the performance of different algorithms. To this end, we contribute a public benchmark framework named ALBench constructed based on our automated deep model training system called YMIR~\citep{YMIR2021}. YMIR could automatically train deep models on given training samples. Based on YMIR, ALBench implements an incremental training pipeline, which first trains initial detection models on a small annotated training set, then iteratively selects samples using active learning models. Selected samples at each step are adopted for updating deep models and active learning models with YMIR. As illustrated in Fig.~\ref{fig:alpipeline}, this procedure is iteratively repeated by ALBench until the stop criterion of active learning is met.

\begin{figure}[t]
\centering
\includegraphics[width=0.8\linewidth]{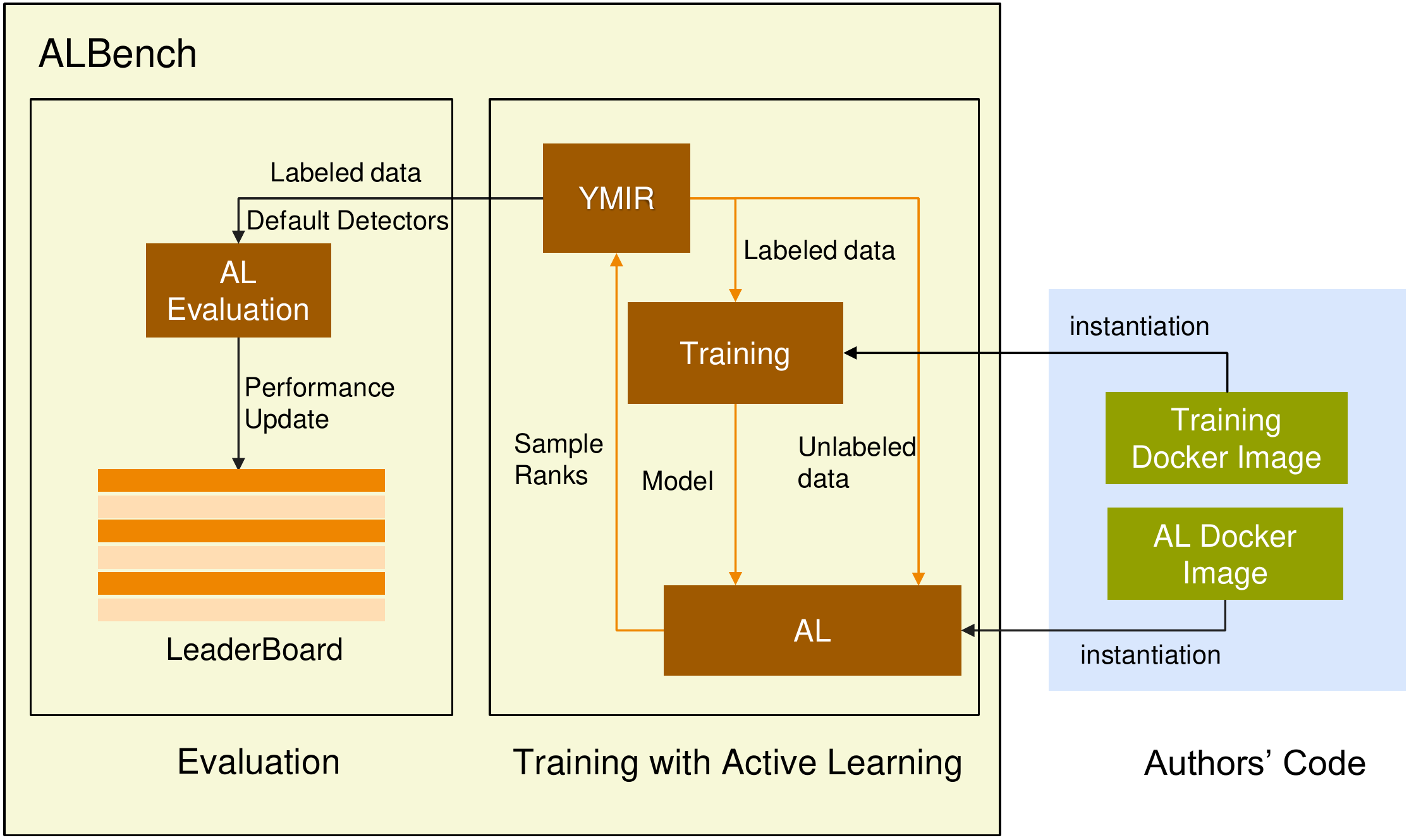}\\
\caption{Illustration about how the authors interact with ALBench as well as detailed process inside ALBench performing the benchmark task.  {\bf Authors' Code:} Implementation provided by authors. {\bf Training with Active Learning:} Model training powered by YMIR with the provided model training code and the active learning code. {\bf Evaluation:} Performance evaluation for the resulting model as well as all intermediate models. Authors could dismiss the "Training Docker Image" part if they employ a standard training methodology that has already been adopted by YMIR. The brown arrows show the looping training cycle.}
\label{fig:albenchoverview}
\end{figure}

ALBench is a plug-and-play framework. The compatibility of various active learning algorithms with ALBench is implemented through docker images which  package active learning models to fit into our training and mining framework. Within ALBench, different active learning algorithms share the same training setup, \emph{e.g.}, training dataset, number of selected samples at each iteration,\emph{ etc.} The subsequent testing stage tests different active learning algorithms with the same setup. We illustrate the training and testing procedures in Fig.~\ref{fig:albenchoverview}, which effectively ensure an objective comparison among different active learning algorithms. Moreover, it is straightforward to study the model behavior using different settings once the related training and active learning docker images are provided. Performance on a setting can be obtained even the original work did not explore the corresponding study. We believe those characteristics of the proposed system will facilitate the research for the active learning community.  


We tested ALBench with its default detector and dataset. Different active learning algorithms including baseline random sampling, entropy~\cite{Roy2018}, ALDD~\cite{Aghdam2019}, and CALD~\cite{Zhu2021} are compared. Besides showing the general comparison, ALBench also illustrates the per-category performance, where many interesting observations can be observed. ALBench is a flexible and allows to involve more testing setups, detectors, and active learning algorithms. It will be publicly released to facility the research community. To the best of our knowledge, this is an original contribution on the benchmark of active learning in object detection.



\section{Background: Active Learning for Object Detection}\label{sec:bkg-detection}
Different active learning algorithms share similar objective of selecting the most valuable training samples. Some algorithms has been adopted to train deep models for image classification. As shown in previous literature~\citep{Caramalau2021}, iteratively selecting training samples saves considerable annotation cost. Object detection is another important vision task, and also suffers from the difficulty of data annotation. Existing active learning works for object detection propose different criteria to select samples. For instance, \citep{Roy2018} proposes a query by committee paradigm and uses the disagreement among convolution layers in the detector backbone to select images. ~\citep{Brust2018} evaluates different aggregation metrics for uncertainty-based active learning methods. \citep{Kao2018} first considers both the classification and localization results of the unlabeled images for sample selection by defining localization tightness and localization stability. \citep{Aghdam2019} calculates the uncertainty of background pixels to select informative samples. \citep{Zhu2021} leverages a consistency-based metric to consider the information of box regression and classification simultaneously. \citep{Yuan2021} treats unlabeled images as instance bags and estimates image uncertainty by re-weighting instances in a multiple instance learning fashion. \citep{Haussmann2020} studies the scalability of active learning, and builds a scalable production system for autonomous driving.

A fair comparison among those algorithms is critical for future studies. As illustrated in Table~\ref{tab:setup-comparison}, current works apply different setups, making a fair comparison difficult. One of related works~\citep{Aghdam2019} focuses on the pedestrian detection task, and is not evaluated on general object detection datasets like COCO~\citep{COCO2014} and VOC~\citep{Everingham2010}. Similarly,~\citep{Di2019} focuses on the object detection in autonomous driving scenarios. Existing approaches also select different numbers of samples at each iteration. For instance,~\citep{Haussmann2020} uses 100k images as the initial training set, and selects 200k images at each iteration. Another work~\citep{Brust2018} trains the YOLO~\citep{Yolo2016} detector and considers the category imbalance issue during training. It only selects 50 images at each iteration. Annotating 200k images at each iteration is expensive for real applications. Selecting too few samples degrades the speed of performance enhancement, leading to a time-consuming training process. Therefore, it not reasonable to select either too many or too few samples for active learning.
 
Existing approaches also follow different testing setups and some approaches suggest contradictory conclusions. Some approaches like~\citep{Yoo2019} only report their performance on the VOC dataset and are not evaluated on the more challenging COCO. A recent work~\citep{Choi2021} trains the SSD~\citep{LiuSSD16} detector and uses both COCO and VOC in experiments. However, it does not report iteration details on COCO and the performance is inferior to commonly used ensemble and dropout methods.~\citet{Kao2018} and~\citet{Zhu2021} provide comprehensive studies on VOC and COCO.~\citet{Kao2018} use random selection as the baseline and the proposed method is only applicable to 2-stage detectors. Both RetinaNet~\citep{Lin2017} and FasterRCNN~\citep{Ren2015} are trained with active learning algorithms proposed by~\citep{Zhu2021}. Experiments in~\citep{Zhu2021} show that the LS/C method proposed in~\citep{Kao2018} is not as good as the random selection baseline, which contradicts with the conclusions in~\citep{Kao2018}.

To summarize, current active learning algorithms for object detection follow different training and testing setups. Some works indicate that, baseline algorithms like random selection perform similarly with more complicated algorithms. It is also interesting to find that, some works draw contradictory conclusions. Those phenomena could be attributed to multiple reasons. Most academic datasets contain high quality samples thanks to their manual annotation and pre-selection procedure, making randomly selected samples present reasonably good quality. Contradictory conclusions can be attributed to different testing and training setups. It is important to build a benchmark platform to 1) Simplify the process to benchmark a method, 2) provide unified training and testing setups, and 3) provide comprehensive evaluations, \emph{e.g.}, show the performance on training and testing sets with different qualities and difficulties. 

\section{Benchmark Framework Design}~\label{sec:design}
This part proceeds to introduce the design and details of ALBench. An overview of ALBench is first presented, followed by detailed introductions on individual components. 

\subsection{Overview}~\label{sec:alb-overview}
Fig.~\ref{fig:albenchoverview} presents the framework of ALBench, which is an automatic training and evaluation system requiring only a few necessary running interfaces from the authors. In ALBench, different active learning algorithms can stick to their own optimal settings and hyper-parameters to conduct sample selection. ALBench incrementally trains object detection models with augmented training sets obtained from the active learning process, and compares their object detection performance.  It provides standard setups for subsequent detection model training and testing to ensure an objective comparison. The final results will be reported on the leaderboard.

ALBench is constructed based on the automated deep model training system called YMIR~\citep{YMIR2021}. YMIR is an open source platform to support the rapid training of deep models at scale. It puts the efficient data development at the center of the machine learning development process, integrates active learning methods, data and model version control, and uses concepts such as projects to enable fast iterations of multiple task specific datasets in parallel. It abstracts the development process into core states and operations, and integrates third party tools via open APIs as implementations of the operations. More details of YMIR can be found in~\citep{YMIR2021}.

As illustrated in Fig.~\ref{fig:albenchoverview}, ALBench is composed of three components including the Input Interface, Training with Active Learning, and Evaluation, respectively.

\subsection{Open Interface}~\label{sec:authorcode}
The open interface (right blue box in Fig.~\ref{fig:albenchoverview}) defines how the users interact with the benchmark system. It defines protocols for packaging active learning algorithms into AL Docker Images, which are hence hosted by YMIR to run sample selection. Without providing detector training code, ALBench has the capability of training the detector with default deep learning models. However, some active learning methodologies are highly coupled with specific detector architectures. To evaluate those active learning algorithms, ALBench allows the users to provide their own object detection training code through a docker image. The docker image will replace the standard training methodology in YMIR to train object detectors and update active learning models. It is fairly simple for using the benchmark system.



\begin{figure}
\centering
\includegraphics[width=0.70\linewidth]{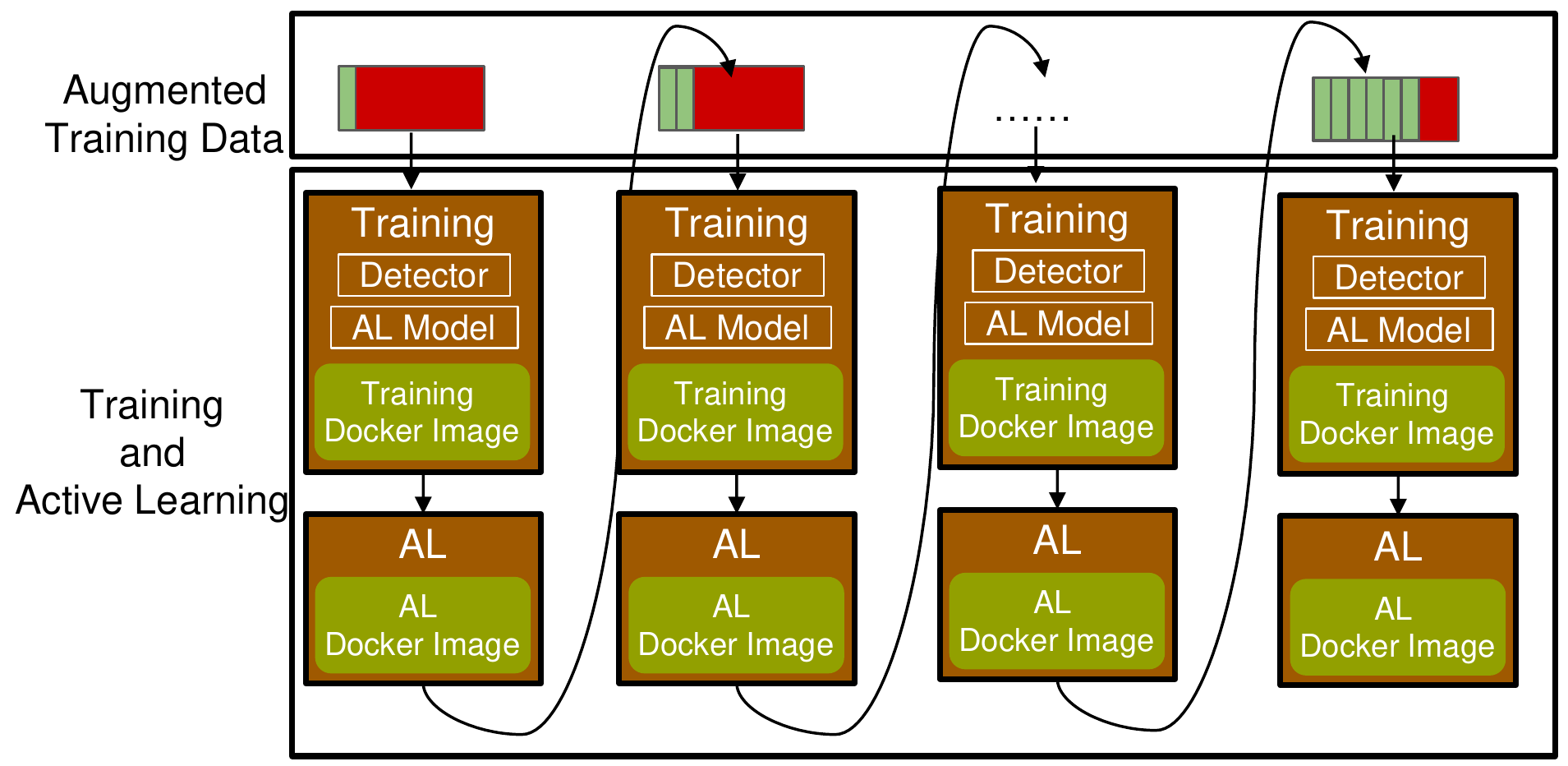}\\
\caption{Illustration of the Training with Active Learning module of ALBench.}
\label{fig:albenchtrainal}
\end{figure}

\subsection{Training with Active Learning}~\label{sec:evaluation}
The Training with Active Learning module takes docker images as input, iteratively runs sample selection, and outputs the augmented training set after each active learning iteration. With docker images as input, the Training with Active Learning module iteratively conducts following steps on a dataset:
\begin{itemize}
    \item Determine new training samples with active learning algorithms
    \vspace{-3mm}
    \item Update object detection models on the augmented training set
    \vspace{-3mm}
    \item Update active learning models with corresponding updated detectors and the provided AL algorithm. 
\end{itemize}

ALBench iteratively repeats above steps till the predefined stop criteria were met. Fig.~\ref{fig:albenchtrainal} illustrates the computational flow chart of the Training with Active Learning module. The training docker image and AL docker image are hosted by the YMIR system. The training docker image is instantiated to train detection models and active learning models. Note that some active learning method requires a training process while others do not. Thus the active learning model training part does not have to be necessary. The AL docker image is instantiated to perform sample selection which gives each input sample a ranking probability for subsequent sample selection. The sampling procedure is implemented by YMIR, simply choosing the top K samples. Different AL algorithms can stick to their optimal settings by packing their code and hyper-parameters in the docker image. ALBench has no dependency on these settings. Meanwhile, all AL methods share identical training setups such as initial training data, number of samples to output for each iteration, stop criteria, \emph{etc.}, to ensure an objective performance comparison.

\subsection{Evaluation}~\label{sec:leaderboard}
\begin{figure}
\centering
\includegraphics[width=0.67\linewidth]{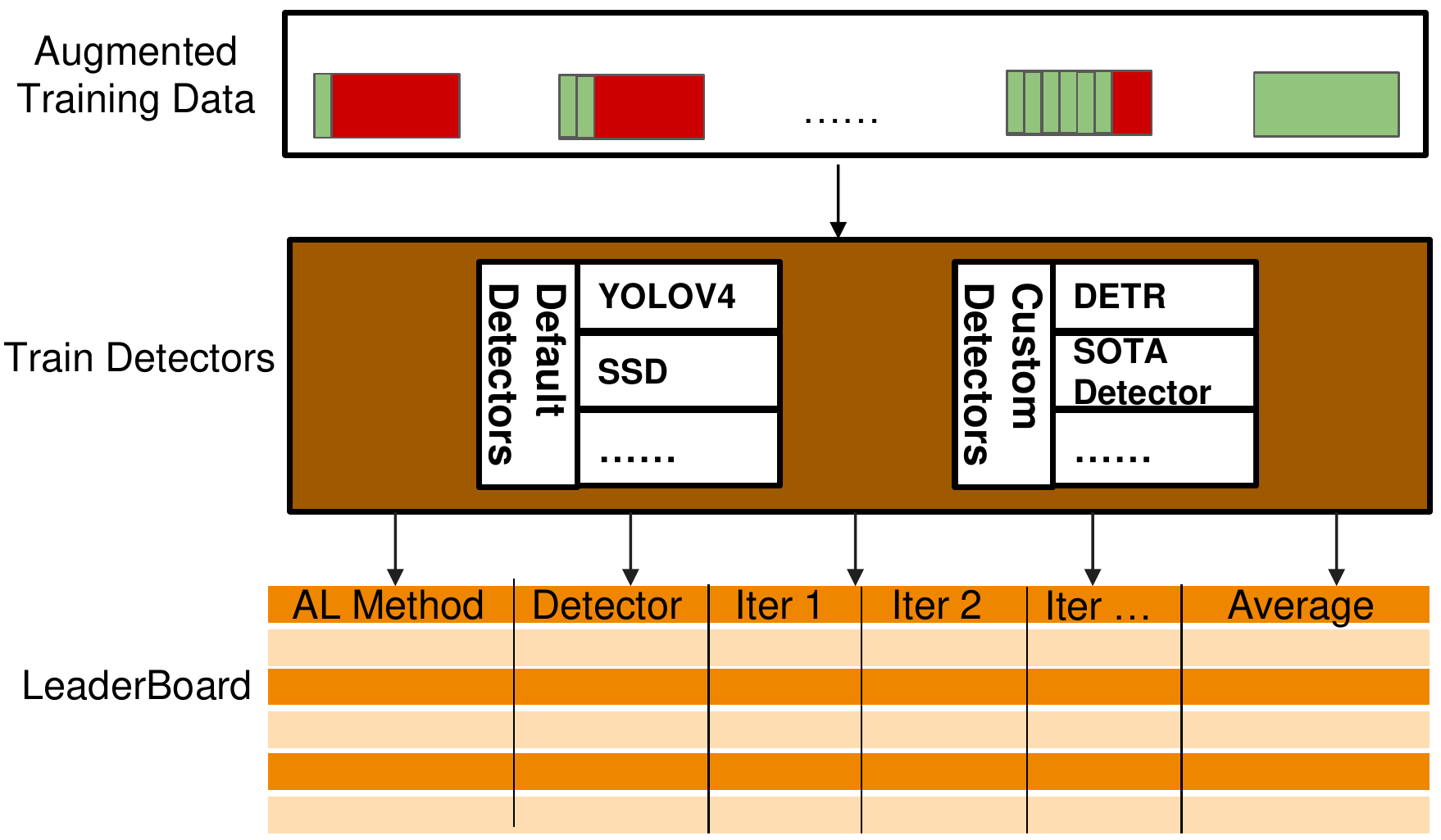}\\
\caption{Illustration of the detailed structure of the automated evaluation module of ALBench.}
\label{fig:albenchevaluate}
\end{figure}

Fig.~\ref{fig:albenchevaluate} shows the detailed structure of the Evaluation module, which provides a unified testing setup for different active learning algorithms. After each training iteration, the Training with Active Learning module outputs the updated training set and updated object detection models to the Evaluation module. The Evaluation module proceeds to evaluate the active learning algorithm by comparing the performance of object detectors trained using their selected samples. To fairly compare the quality of their selected samples, the Evaluation module will also train default object detectors with intermediate augmented training sets in each iteration. Some active learning algorithms are coupled with specifically designed detectors and training algorithms. Besides training those detectors with standard methodology provided by ALBench, the Evaluation module also trains them with their specific training algorithms provided by users. In other words, some detectors are trained by both commonly followed methodology and their specific ones to distinguish the contribution from sample selection with active learning from detectors themselves.

\begin{figure}
\centering
\includegraphics[width=0.7\linewidth]{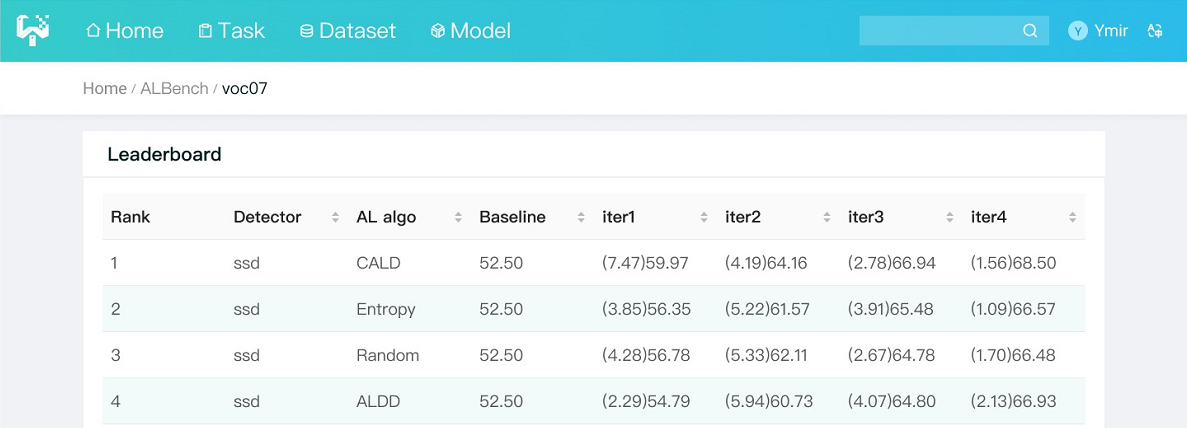}\\
\caption{Illustration of the leaderboard.}
\label{fig:leaderboard}
\end{figure}

The performance evaluation is conducted after each active learning iteration and is reported on a leaderboard as shown in Fig.~\ref{fig:leaderboard}. Note that, ALBench allows to flexibly add new testing setups and state-of-the art object detectors. Different testing setups and detectors can be adjusted or added through a user interface.

\section{Benchmark Showcase}~\label{sec:experiment}
This section compares different active learning algorithms for object detection with ALBench. It first introduces training and testing setups, then shows the performance comparison, followed by discussions.

\subsection{Default Setups}
With the automated training pipeline, training and testing setups can be flexibly defined in ALBench, \emph{e.g}., to include new datasets and train new detectors. This paper simply tests ALBench with default setups to show its validity. 

\textbf{Datasets:} ALBench uses two widely used object detection datasets Pascal VOC 2007 and VOC 2012~\citep{Everingham2010} as the default training set. For VOC2007, we use its trainval set for training and sample mining, and test on its testset. On VOC2012, we use its train set for training and sample mining, and test on the validation set. We randomly select 1000 images as the initial training set, and sample 1000 images in each iteration. The training stops after 4 active learning iterations. 

\textbf{Algorithms:} 
ALBench adopts the SSD300~\citep{LiuSSD16} implemented with Vgg16~\citep{Simonyan2015} as the default object detection model. Codes for training are acquired from the mxnet model zoo. Each active learning iteration trains for 100 epochs and picks the best model with highest mAP on the testset for sample selection. Default active learning algorithms include two commonly used baselines, \emph{i.e}., Entropy~\citep{Roy2018} and Random Selection~\citep{Choi2021}, and two recent approaches ALDD~\cite{Aghdam2019} and CALD~\cite{Zhu2021}, respectively. 

\begin{table}[] 
\centering
\footnotesize
\begin{tabular}{c|c|c|c|c|c|c}
\hline
Dataset    & Method   & baseline & iter1 & iter2 & iter3 & iter4 \\ \hline
\multirow{4}{*}{VOC07}
& SSD+ALDD & 52.50     & (+2.29) 54.79    & (+5.94) 60.73     & (+4.07) 64.80    & (+2.13) 66.93    \\ \cline{2-7}
& SSD+CALD & 52.50     & (+7.47) {\bf 59.97}    & (+4.19) {\bf 64.16}     & (+2.78) {\bf 66.94}    & (+1.56) {\bf 68.50}    \\ \cline{2-7}
& SSD+RANDOM  & 52.50  & (+4.28) 56.78    & (+5.33) 62.11     & (+2.67) 64.78    & (+1.70) 66.48     \\ \cline{2-7}
& SSD+ENTROPY & 52.50         & (+3.85) 56.35    & (+5.22) 61.57     & (+3.91) 65.48     & (+1.09) 66.57     \\ \hline\hline

\multirow{4}{*}{VOC12}
&SSD+ALDD & 48.29  & (+3.96) 52.25     & (+5.20) 57.45     & (+2.84) 60.29    & (+2.19) 62.48     \\ \cline{2-7}
&SSD+CALD & 48.29  & (+7.64) \bf 55.93     & (+2.28) \bf 58.21     & (+3.36) \bf 61.57    & (+0.94) \bf 62.51    \\ \cline{2-7}
&SSD+RANDOM  & 48.29  & (+5.19) 53.48    & (+4.53) 58.01     & (+2.35) 60.36    & (+1.73) 62.09     \\ \cline{2-7}
&SSD+ENTROPY & 48.29  & (+5.68) 53.97    & (+3.50) 57.47     & (+3.18) 60.65     & (+1.80) 62.24    \\ \hline
\end{tabular}
\caption{Performance of AL methods evaluated through our system}
\label{tab:perf-comparison}
\end{table}

\subsection{Results and Discussions} 
Table~\ref{tab:perf-comparison} presents the performance comparison among default active learning algorithms in ALBench. The recent CALD outperforms ALDD and baselines on VOC07 and VOC12. At iter1, CLAD substantial outperforms others. As more training samples are selected, those AL algorithms start to show similar performance. This is because the training sets of VOC07 and VOC12 only contain about 5K and 5.7K images, respectively, which minish the importance and contribution of AL algorithms as more samples are selected. Another interesting observation is that ALDD does not perform as good as simple baseline AL algorithms at iter1 and iter2. This could be because VOC training set generally contains high quality samples, making random selection achieves reasonable good performance as claimed in many studies~\cite{Zhu2021}.

\begin{figure}
\centering
\includegraphics[width=0.95\linewidth]{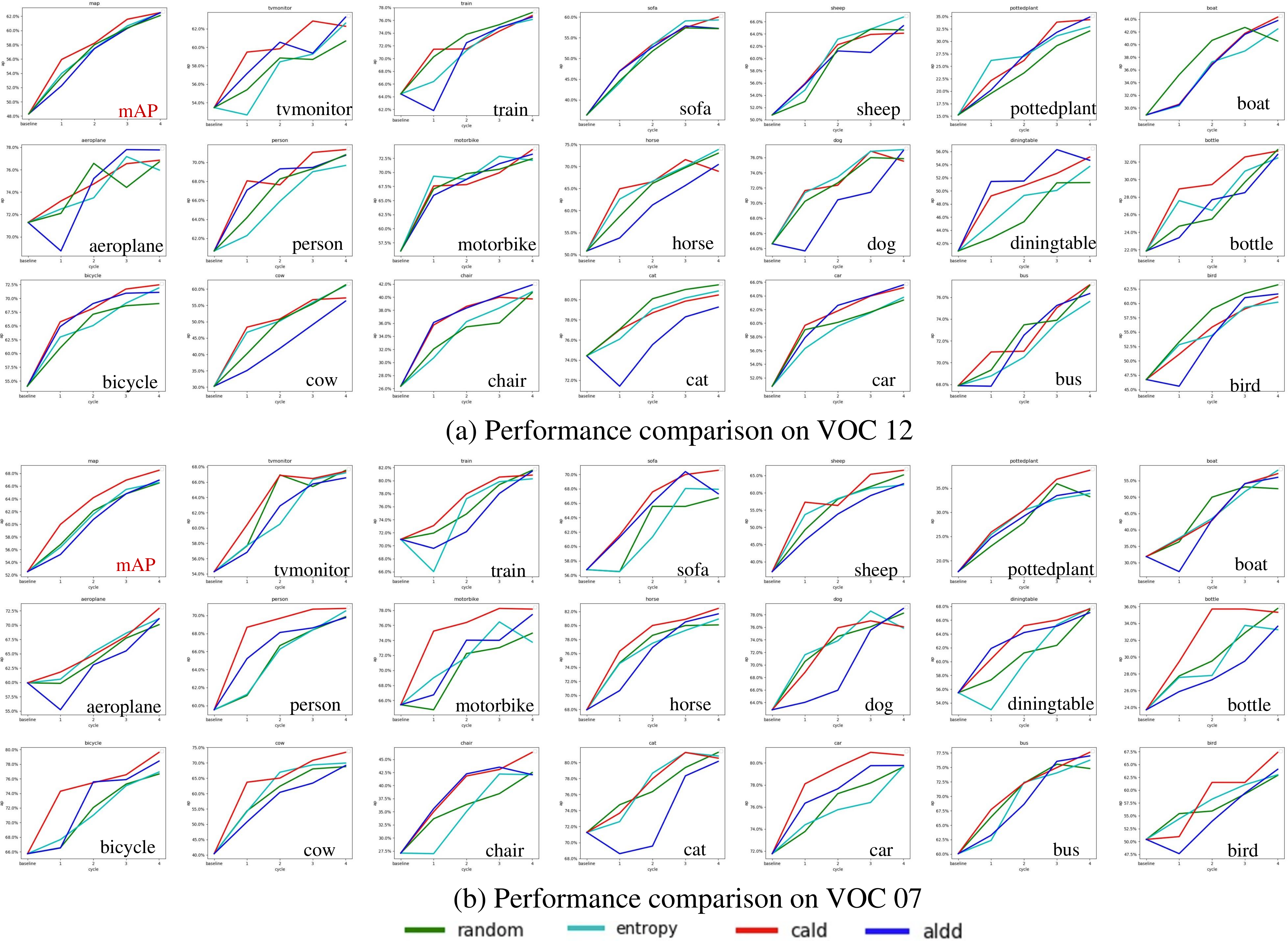}\\
\caption{Illustration of a detailed per-category performance comparison.}
\label{fig:visualizedcompare}
\end{figure}

ALBench could also show a detailed per-category performance comparison as illustrated in Fig.~\ref{fig:visualizedcompare}, where CALD does not always perform the best. For example, on the horse, cat, and sheet categories of VOC12 as well as the dog category of VOC07, CALD performs worse than random selection. In Fig.~\ref{fig:visualizedcompare}, many AL algorithms are not stable, \emph{e.g.}, substantial performance degradation occurs for CALD and ALDD as more samples are selected on many categories. Fig.~\ref{fig:visualizedcompare} also indicates that, low baseline detector performance does not substantially degrade the effectiveness of subsequent sample selection on VOC07 and VOC12. For instance, the baseline detector only achieves about 26\% AP on chair category. AL algorithms effectively boost the AP by about 16\% on both datasets. This could also be attributed to the fact that VOC datasets have been dedicatedly prepared with sample selection and noise filtering, which simplifies high quality sample selection. Using realistic datasets containing larger portions of noises, redundancies, and long-tailed catregories has potential to substantially evaluate the value and performance of active learning algorithms.


\section{Limitation and Broader Impact}~\label{sec:limitation}
 ALBench currently focuses on object detection task. Further work will be conducted to extend ALBench to more vision and machine learning tasks. Besides default setups, more active learning algorithms, larger and realistic training and testing sets will be included in ALBench to make more comprehensive comparisons. As the first open source benchmark platform for active learning, ALBench is expected to bring broad impact to the field of active learning.

\section{Conclusions}~\label{sec:conclusion}
This paper proposes an active learning benchmark framework named as ALBench to simplify the implementation and comparison of active learning algorithms in object detection. ALBench is powered by an automatic deep model training system, which applies unified training and testing setups for different active learning algorithms. ALBench framework is easy-to-use, compatible with different active learning algorithms, and ensures their objective comparison. ALBench also allows to flexibly add more training and testing setups, and detectors. As an original contribution on the benchmark of active learning, ALBench is expected to benefit the active learning community. 

%

\bibliography{egbib}

\end{document}